\documentclass{article}
\usepackage{spconf,amsmath,graphicx}
\usepackage{multirow}
\usepackage[table,xcdraw]{xcolor}
\usepackage{hyperref}
\usepackage{soul}
\usepackage{blindtext}

\title{Lightweight high-resolution Subject Matting in the Real World}
\name{Peng Liu, Fanyi Wang, Jingwen Su, Yanhao Zhang, Guojun Qi}
\address{OPPO Research Institute}
\begin{document}
%
\maketitle
\begin{abstract}
Existing saliency object detection (SOD) methods struggle to satisfy fast inference and accurate results simultaneously in high resolution scenes. 
They are limited by the quality of public datasets and efficient network modules for high-resolution images. 
To alleviate these issues, we propose to construct a saliency object matting dataset HRSOM and a lightweight network PSUNet. Considering efficient inference of mobile depolyment framework, we design a symmetric pixel shuffle module and a lightweight module TRSU.
Compared to 13 SOD methods, the proposed PSUNet has the best objective performance on the high-resolution benchmark dataset.
Evaluation results of objective assessment are superior compared to U$^2$Net that has 10 times of parameter amount of our network.
On Snapdragon 8 Gen 2 Mobile Platform, inference a single 640$\times$640 image only takes 113ms. And on the subjective assessment, evaluation results are better than the industry benchmark IOS16 (Lift subject from background).

\end{abstract}
\section{Introduction}
\label{sec:intro}
SOD is a technique for automatic identification and localization of the most salient objects.
SOD algorithms can be classified into traditional methods and deep learning methods.
Traditional methods mainly use low-level features such as color, brightness, orientation, and contrast of an image to compute saliency maps, but they often fail to deal with complex background and multiple objects \cite{yang2013saliency,zhu2014saliency,srivatsa2015salient}.
Deep learning methods utilize CNNs to learn high-level semantic features to improve performance of SOD.
One of the challenges of SOD is to extract saliency maps with clear boundaries from high-resolution images while controlling the computational cost. 
Practical SOD algorithms need to balance between speed and accuracy, in order to adapt to various application requirements and hardware conditions.
In this regard, Wu et al.\cite{wu2021mobilesal}, Lee et al.\cite{lee2021tracer}, Wang et al.\cite{wang2021raining} designed lightweight structures to accelerate inference.
Liu et al.\cite{Liu2019PoolSal}, Wu et al.\cite{Wu_2022}, Qin et al.\cite{Qin_2020_PR, qin2022} used multi-scale feature fusion to improve the quality of saliency maps.
Qin et al.\cite{qin2021boundary} used coarse-to-fine framework to obtain finer results. 
Xie et al.\cite{xie2022pyramid} , Kim et al.\cite{kim2022revisiting} designed high resolution modules to obtain accurate high resolution saliency maps.

High-resolution SOD datasets \cite{7780454, xie2022pyramid, zeng2019highresolution} also contribute to the advancement of SOD algorithms.
They contain images of better quality and labels of higher accuracy, compared to low-resolution SOD datasets. 
Most existing SOD methods can achieve good performances on public datasets, but they are unable to apply on real scenes. 
Firstly, most SOD datasets are low-resolution, which hinders the algorithm to process high-resolution images. Secondly, binary labels define SOD as a binary segmentation task of foreground and background. And label edges of some datasets have jaggedness, leading to difficulty for the algorithm to derive fine results. 
Lastly, saliency labels have large deviation from subjective perception, which is not friendly for learning. 



Our contributions are summarised as following.
We propose a lightweight SOD network PSUNet, which is designed with full considerations of mobile adaptation to achieve a good balance between high resolution and efficiency.
We propose a symmetric pixel shuffle module to effectively enhance high-resolution SOD results, at a very low cost of computation and parameter amount increment.
We construct a saliency object matting dataset HRSOM consisting of 13w+ high-quality and high-resolution images, based on almost all available saliency detection datasets and self-collected datasets.
We design a set of training methods and subjective and objective measures for real-world datasets, enabling the algorithm to robustly cope with open-world problems.

\begin{figure*}[t]
\centering
\includegraphics[width=0.82\linewidth]{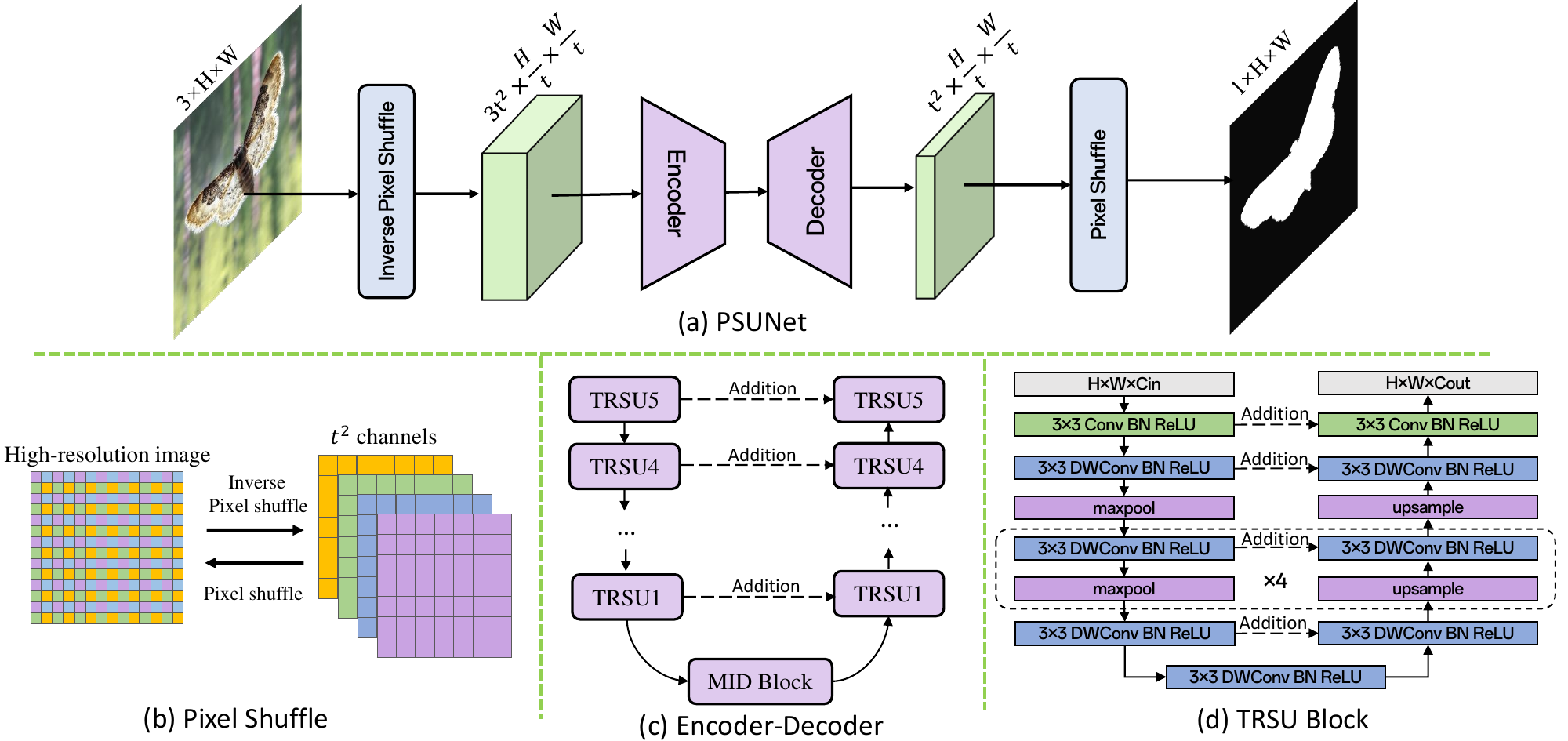}
\caption{Overall structure of PSUNet.}
\label{figure2}
\end{figure*}
\section{Method}
\subsection{PSUNet}
Our proposed PSUNet adopts the UNet \cite{ronneberger2015unet} architecture and a symmetric pixel shuffle module (SPSM) for high-resolution scenarios. It greatly improves the capability of processing high-resolution images with only tiny increment of computation and parameter amount.
Due to limited computation power on mobile devices, network width has much larger impact on inference speed than depth. Hence during the design of base module, we attempt to control the network depth as much as possible while reducing computation cost of single convolution. And we design a Tiny Residual Ublock (TRSU). To control the network width, skip strategy takes the form of summation of feature maps.
\subsubsection{Symmetrical Pixel Shuffle Module}


PixelShuffle \cite{shi2016real} is widely used in image super-resolution tasks. It is a very efficient sampling method to obtain high-resolution feature maps from low-resolution ones, through convolution and reorganization of multiple channels. 
Inverse pixel shuffle process extracts high resolution image features and preserves global features. For SOD task, global information is the key to detect salient objects. At the same time, high-resolution input also retains more detailed information. Compared to bilinear interpolation for downsampling, inverse pixel shuffle has no loss on details, as demonstrated in the later section of ablation study.
Implementation of Symmetrical Pixel Shuffle Module is shown in Fig. \ref{figure2}(b). 
In the upsampling process, a grid of t$\times$t is constructed for each low-resolution pixel. Feature values at the corresponding positions in t$\times$t feature maps are utilised to fill into these small grids according to predefined rules. 
And downsampling process conducts reorganization by following the same rules and fills the small grids divided by each low-resolution pixel. 
In this process, the model can adjust weights of the t$\times$t shuffle channels and continuously optimizes the generated results, which can be expressed as below,
\begin{eqnarray}
\small
N(C\times t\times t)HW & = & NC(H\times t)(W\times t)
\end{eqnarray}
where $N$ represents number of batches, $C$ represents number of channels, $H$ and $W$ represent height and width of feature map, $t$ represents sampling ratio.


\subsubsection{Encoder-Decoder}
The overall structure of Encoder-Decoder is similar to U$^2$Net \cite{Qin_2020_PR} and consists of the proposed lightweight TRSU (Tiny ReSidual Ublock) module.
As shown in Fig. \ref{figure2}(d), the proposed TRSU module also has the structure of UNet. After a layer of 3$\times$3 convolution, feature maps are sampled to a smaller size by a stack of multiple combinations of depth separable convolution and max pooling. 
Then after several layers of depth separable convolution, resolution is restored by the same number of combinations of upsampling and depth separable convolution. 
The final output is derived after a layer of 3$\times$3 convolution.
Skip connections in TRSU module are in the form of addition to control the network width.
MIDBLock (Fig. \ref{figure2}(c)) stacks pure depth-separable convolution structures to extract deeper features. For TRSU module in the decoder, all convolutions in Fig. \ref{figure2}(d) are replaced with depth-separable convolutions. Finally, feature maps are fed into SPSM after a single layer of convolution.
Skip connections between the encoder and the decoder are also in the form of addition.

\subsection{HRSOM Dataset}
\begin{figure}[t]
\centering
\includegraphics[width=0.48\textwidth]{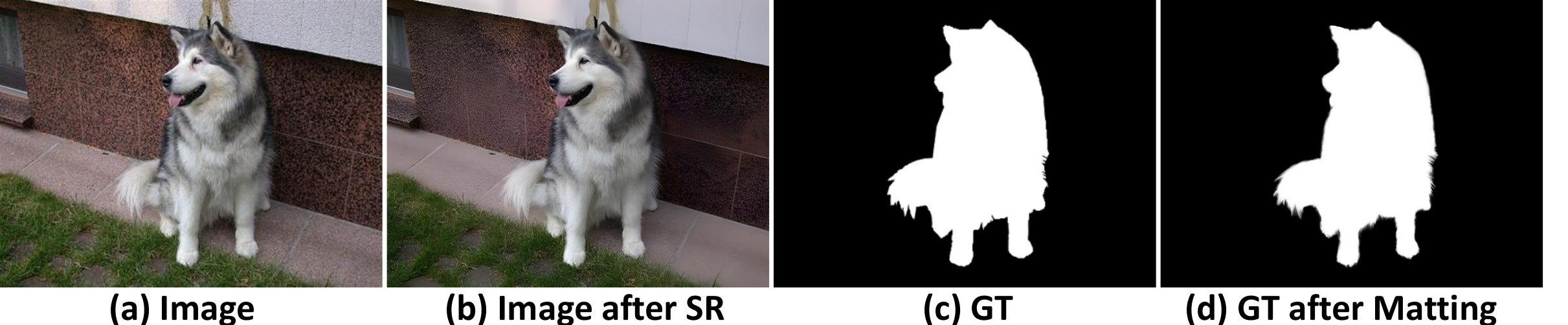}
\caption{Image and label before and after processing.}
\label{SRMatting}
\end{figure}

We integrated RGB SOD low-resolution datasets DUTS-TR, HKU-IS, RGB SOD high-resolution datasets UHRSD, HRSOD, DIS5K, RGB-D SOD datasets COME15K \cite{cascaded_rgbd_sod}, and matting datasets AM, P3M. To construct the integrated dataset, we collaborated with data label experts to establish standards of saliency criteria for practical applications. 
And we manually excluded images unsuitable for labeling, especially for scenarios where multiple datasets behave inconsistently. 
Real super-resolution (SR) algorithm Real-ESRGAN \cite{Wang_2021_ICCV} was used to increase resolution to more than 720 pixels on the short side. Bilinear interpolation for upsampling was used to increase label resolution by the same rate. Labels of all SOD datasets are refined using the pre-trained large matting model. Fig. \ref{SRMatting} shows the results of low resolution images before and after processing. It can be seen that the image details are well preserved after SR, and edge details of label results after matting are significantly improved.
Collected 6w+ data through self-acquisition and purchase are manually labeled according to the established saliency criteria. 
Public data after cleaning and processing and self-collected data after labeling are integrated into the HRSOM dataset of 13w+ data in total.

\subsection{Real scenes Training Strategy}
\subsubsection{Hybrid Loss}
Considering three granularity levels of pixel, image block and full image, we propose a hybrid Loss (Eq.\ref{eq:hybrid}.) consisting of BCE Loss, SSIM loss and IOU loss. 
BCE Loss (Eq.\ref{eq:bce}) is a pixel-level loss function that compares the predicted probability of each pixel with the ground truth label, and computes the negative log-likelihood as results. It allows the model to learn whether each pixel belongs to the salient object or not.
\begin{equation}
\small
\begin{split}
    \mathcal{L} _{bce} &=-\sum_{i=1}^{H} \sum_{j=1}^{W} [G_{ij}\log_{}{(P_{ij})}+ (1-G_{ij})\log_{}{(1-P_{ij})} ]
\end{split}
\label{eq:bce}
\end{equation}
SSIM loss \cite{wang2004image} characterizes structural similarity and is of image block level. It calculates the differences between two image blocks in terms of luminance, contrast, and structural similarity. It enables the model to learn local structural features of salient objects, especially the edges.
IOU loss is an image-level loss function that calculates the degree of overlap between two saliency maps. It allows the model to learn the overall shape and positional information of salient object.
\begin{equation}
\small
\label{eq:iou}
\mathcal{L}_{iou}=1-\frac{\sum_{i=1}^{H} \sum_{j=1}^{W}P_{ij}G_{ij}  }{\sum_{i=1}^{H} \sum_{j=1}^{W}(P_{ij}+G_{ij}-P_{ij}G_{ij})}  
\end{equation}
Combination of these three loss functions can achieve pixel level supervision on both global semantics and local details.
\begin{equation}
\small
\label{eq:hybrid}
\mathcal{L} _{total}=  \mathcal{L}_{bce}+\mathcal{L}_{ssim}+  \mathcal{L}_{iou}
\end{equation}

\subsubsection{Multi-Stage Training}
For the open-world scenario, we design a two-stage training strategy.
In the first stage, a traditional SOD training approach is used to improve detection accuracy of salient objects. This process uses hybrid Loss for training to derive high maxF Score. Similar to most lightweight SOD methods, resultant saliency maps have artifacts and normal edges, leading to unsatisfactory subjective effects.
In the second stage, a matting training approach is adopted to fine-tune the model trained in the first stage. A combination of L1 loss and edge loss is used, and learning rate is reduced to 0.3 times (i.e. 0.0003) of the first stage. After this stage of training, while maintaining saliency object detection accuracy, quality of edges in our results are enhanced, artifacts problems are effectively circumvented, and subjective results are significantly improved.
\section{Experiments}
\label{sec:exps}
\subsection{Implementation Details}
Training batch size is set to 16.  Adam optimizer \cite{kingma2017adam} is used to train our network and its hyper parameters are set to the default values. 
Initial learning rate is set to 1e-3, betas are set to (0.9, 0.999), epsilon 1e-8, weight decay set to 0. Models are trained on 8 V100 GPUs until convergence. 
For data augmentation, images are resized to 680$\times$680, and then randomly cropped to 640$\times$640. In experiments, we add random horizontal flip, random resize small with padding, and random noise. For test, input images are resized to 640$\times$640 for inference to obtain saliency maps. Then saliency maps are resized to the original input image size. Bilinear interpolation is adopted for resizing. Three common metrics MAE, maxF and S-measure are adopted to evaluate our model and SOTA models.
\subsubsection{Datasets}
For fair comparison, the PSUNet is trained on DUTS-TR dataset \cite{8099887}. 
Validation is conducted on three datasets, DAVIS-S \cite{7780454}, HRSOD-TE \cite{zeng2019highresolution}, and UHRSD-TE \cite{xie2022pyramid}. DAVIS-S has 92 high-resolution images from the DAVIS dataset. 
HRSOD dataset is designed for high-resolution SOD, containing 1610 training images and 400 test images.
UHRSD dataset is the highest resolution SOD dataset available, containing 5920 4K-8K resolution images, 4932 training images and 988 test images. And we use its downscaled version of 2K resolution.
Due to low resolution of DUTS-TR and our high resolution structure design, training with only DUTS-TR tend to affect the high resolution image results. Some high resolution methods also train on HRSOD-TR and UHRSD-TR dataset. We follow the same training scheme to compare with the corresponding methods. Besides, our training is also conducted on DUTS-TR after super-resolution.






\subsection{Comparison experiments}
\subsubsection{Comparison with the SOTAs}
We compare the proposed PSUNet with 13 other lightweight SOD methods. For fair comparison, model parameter amount, computational capacity, and saliency maps are either from open-source code or provided by the authors. Results in Table \ref{Comparison with the SOTAs} show that we outperform most of the lightweight methods on three high-resolution datasets, with much smaller computation capacity and parameter amount. 
The proposed PSUNet contains 4.15M parameters and requires 12.38G MACs of computation when the input image size is 640$\times$640. It is much lighter than the majority of SOD methods. Experiment results show that the objective indicators are even comparable to U$^2$Net.


\begin{table}[]
\resizebox{1.0\linewidth}{!}{%
\begin{tabular}{lll|lll|lll|lll}
\hline
\multirow{2}{*}{Method} &
  \multicolumn{1}{l}{\multirow{2}{*}{MACs}} &
  \multicolumn{1}{l|}{\multirow{2}{*}{Params}} &
  \multicolumn{3}{c|}{DAVIS-S} &
  \multicolumn{3}{c|}{HRSOD-TE} &
  \multicolumn{3}{c}{UHRSD-TE} \\ \cline{4-12} 
         & \multicolumn{1}{l}{} & \multicolumn{1}{l|}{} & maxF↑       & MAE↓  & Sm↑   & maxF↑       & MAE↓  & Sm↑   & maxF↑       & MAE↓  & Sm↑   \\ \hline
BASNet                   & 127.04                                     & 87.06                                         & .863                        & .039                        & .881                        & .881                              & .038                        & .890                        & .901                        & .053                        & .883                        \\
PoolNet                  & 89.00                                         & 68.21                                         & {\ul .914}                  & .022                        & .919                        & {\ul .888}                        & .041                        & .895                        & {\ul .908}                   & .052                        & .888                        \\
EGNet                    & 120.15                                     & 111.69                                        & {\ul .913}                  & .023                        & .922                        & {\ul .894}                        & .038                        & .899                        & {\ul .916}                  & .049                        & .895                        \\
SCRN                     & 15.10                                       & 25.23                                         & {\ul .807}                  & .024                        & .912                        & {\ul .896}                        & .034                        & .904                        & {\ul .918}                  & .048                        & .895                        \\
F3Net                    & 16.43                                      & 25.54                                         & {\ul .918}                  & .020                        & .913                        & {\ul .894}                        & .035                        & .897                        & {\ul .916}                  & .045                        & .891                        \\
MINet                    & 86.92                                      & 162.38                                        & {\ul .896}                  & .024                        & .904                        & {\ul .903}                        & .034                        & .900                        & {\ul .920}                  & .043                        & .894                        \\
LDF                      & 15.51                                      & 25.15                                         & .916                        & .019                        & .922                        & .911                              & {\color[HTML]{FE0000} .032} & .905                        & .919                        & .047                        & .890                        \\
GateNet                  & 161.91                                     & 128.63                                        & {\ul .917}                  & .024                        & .915                        & {\ul .912}                        & {\color[HTML]{FE0000} .032} & {\color[HTML]{FE0000} .911} & {\ul .921}                  & .048                        & .895                        \\
PFSNet                   & 37.39                                      & 31.18                                         & {\ul .910}                  & .019                        & .922                        & {\color[HTML]{3166FF} {\ul .914}} & {\color[HTML]{FE0000} .032} & {\color[HTML]{3166FF} .907} & {\ul .920}                  & .042                        & .899                        \\
CTDNet                   & {\color[HTML]{FE0000} 10.15}               & 24.63                                         & {\ul .917}                  & .021                        & .911                        & {\ul .907}                        & .033                        & .898                        & {\ul .917}                  & .050                        & .883                        \\
ICON                     & \multicolumn{1}{l}{64.90}                  & \multicolumn{1}{l|}{19.17}                    & \multicolumn{1}{l}{.923}    & \multicolumn{1}{l}{.023}    & \multicolumn{1}{l|}{.914}   & \multicolumn{1}{l}{.908}          & \multicolumn{1}{l}{.041}    & \multicolumn{1}{l|}{.894}   & \multicolumn{1}{l}{.913}    & \multicolumn{1}{l}{.050}    & \multicolumn{1}{l}{.888}    \\
U2Netp                   & 19.80                                       & {\color[HTML]{FE0000} 1.13}                   & .889                        & .038                        & .891                        & .886                              & .046                        & .883                        & .915                        & .059                        & .878                        \\
U2Net                    & \multicolumn{1}{l}{58.66}                  & \multicolumn{1}{l|}{44.01}                    & \multicolumn{1}{l}{.911}    & \multicolumn{1}{l}{.026}    & \multicolumn{1}{l|}{.911}   & \multicolumn{1}{l}{.900}          & \multicolumn{1}{l}{.039}    & \multicolumn{1}{l|}{.897}   & \multicolumn{1}{l}{.917}    & \multicolumn{1}{l}{.049}    & \multicolumn{1}{l}{.889}    \\ \hline
Ours-D                   & {\color[HTML]{333333} 12.38}               & {\color[HTML]{000000} 4.15}                   & .936                        & .018                        & .926                        & .888                              & .042                        & .887                        & .909                        & .054                        & .884                        \\
Ours-DH                  & 12.38                                      & 4.15                                          & {\color[HTML]{3166FF} .952}                        & {\color[HTML]{3166FF} .014} & {\color[HTML]{3166FF} .938} & .907                              & .039                        & .898                        & .922                        & .046                        & .897                        \\
Ours-HU                  & 12.38                                      & 4.15                                          & .943                        & .017                        & .934                        & .900                              & .043                        & .888                        & .922                        & .048                        & .891                        \\
Ours-DHU                 & 12.38                                      & 4.15                                          & .946 & .016                        & .936                        & .908                              & .038                        & .896                        & {\color[HTML]{3166FF} .942} & {\color[HTML]{FE0000} .035} & {\color[HTML]{FE0000} .918} \\
Ours-ALL                 & {\color[HTML]{3166FF} 12.38}               & {\color[HTML]{3531FF} 4.15}                   & {\color[HTML]{FE0000} .958} & {\color[HTML]{FE0000} .012} & {\color[HTML]{FE0000} .948} & {\color[HTML]{FE0000} .922}       & .041                        & .896                        & {\color[HTML]{FE0000} .948} & {\color[HTML]{3531FF} .038} & {\color[HTML]{3166FF} .915} \\ \hline
\end{tabular}}
\caption{Objective comparison with the SOTAs. D represents DUTS-TR, H represents HRSOD-TR, U represents UHRSD-TR, and ALL represents our integrated HRSOM dataset. The best results are marked in red, second is in blue.}
\label{Comparison with the SOTAs}
\end{table}

\begin{table}[t]
\centering
\resizebox{0.8\linewidth}{!}{%
\begin{tabular}{l|ll|ll}

\hline
& \multicolumn{2}{l|}{1st percentage↑} & \multicolumn{2}{l}{3rd percentage↓} \\ \cline{2-5} 
\multirow{-2}{*}{Dimension} & Ours              & IOS             & Ours             & IOS             \\ \hline
Subject Completeness & {\color[HTML]{000000} \textbf{35\%}} & {\color[HTML]{000000} 33\%} & {\color[HTML]{000000} \textbf{25\%}} & {\color[HTML]{000000} 49\%} \\
Edge Accuracy     & {\color[HTML]{000000} \textbf{36\%}} & {\color[HTML]{000000} 28\%} & {\color[HTML]{000000} \textbf{25\%}} & {\color[HTML]{000000} 55\%} \\ \hline
\end{tabular}}
\caption{Subjective comparison with IOS based on user study.}
\label{tab:Comparison with IOS based on user study}
\end{table}

\begin{figure}[t]
\centering
\includegraphics[width=0.5\textwidth]{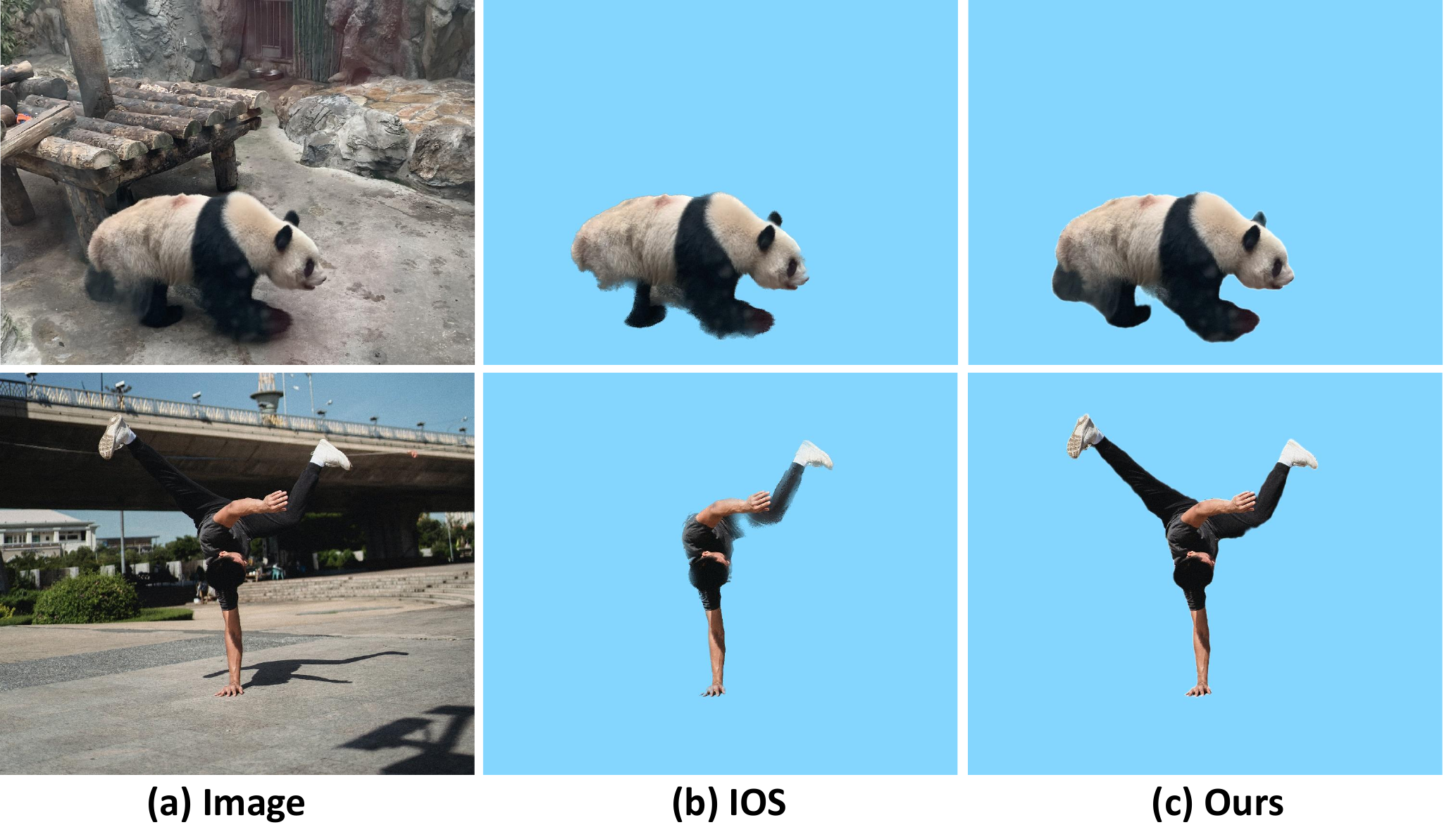}
\caption{Visual comparison with IOS}
\label{vs-ios}
\end{figure}
\begin{table}[t]
    \centering
    \resizebox{0.8\linewidth}{!}{%
    \begin{tabular}{l|l|lll}
    \hline
    Ablation     & Configurations &maxF↑ &MAE↓ & Sm↑ \\ \hline
    \multirow{4}{*}{Arch.} & No SPSM                 & 0.905          & 0.059         & 0.877        \\
                           & SPSM with t=2           & 0.922          & 0.046         & 0.897        \\
                           & SPSM with t=3           & 0.937          & 0.042         & 0.907        \\
                           & SPSM with t=4           & \textbf{0.940}          & \textbf{0.040}         & \textbf{0.909}        \\ \hline
    \multirow{4}{*}{Loss}  & BCE                     & 0.918          & 0.057         & 0.886        \\ 
                           & BCE + SSIM              & 0.921          & 0.053         & 0.890        \\ 
                           & BCE + IoU               & 0.918          & 0.050         & 0.886        \\ 
                           & BCE + SSIM + IoU        & \textbf{0.922}          & \textbf{0.046}         & \textbf{0.897}        \\ \hline
    \end{tabular}}
    \caption{Ablation Study}
    \label{Ablation Study}
\end{table}
\subsubsection{Subjective evaluation}
We obtain robust open-world lightweight SOD model based on the proposed HRSOM dataset with two-stage training strategy. After float16 quantization, they are deployed to mobile devices. Model inference takes only 113ms on the high-performance Snapdragon 8 Gen 2 Mobile Platform, and 234ms on the middle level chip platform MediaTek Dimensity 8100-Max.
To compare with industrial SOTA products, we collected 500 open-world images to derive the IOS16 Quick Matting results and our results. These results were shuffled and removed labels. At each time, 100 sets of images were randomly selected and present to 93 experienced subjective reviewers for blind evaluation and scoring, according to subject completeness and edge accuracy.
For fairness, one result from our earlier trained models was added to make up a group of 3 images per set. The results of the subjective scoring ranking are shown in Table \ref{tab:Comparison with IOS based on user study}, and some hard case comparison results are shown in Fig. \ref{vs-ios}.

\subsection{Ablation study}

In this section, we study the effect of different sampling ratio t of SPSM on objective metrics, and verify the effectiveness of hybrid loss.
In Table \ref{Ablation Study}, we compare the difference in results with and without SPSM, and results at different values of sampling ratio t in SPSM. Models are trained on HRSOD-TR and UHRSO-TR datasets. Experiments demonstrate that our SPSM has higher improvement in high resolution scenes compared to no SPSM. And for scenes with large resolution, the larger value of t, the more significant improvement.
We also compare effects of different combinations of loss functions. SPSM is set with t=2 and size of training image 640$\times$640. Models were trained on HRSOD-TR and UHRSD-TR dataset and tested on UHRSD-TE dataset. Experiment results prove that our Hybrid Loss is meaningful.

\section{Conclusions}
In this paper, we propose a lightweight salient object matting network PSUNet for high-resolution images. Compared with 13 SOD SOTA methods on three high-resolution SOD datasets, the proposed PSUNet achieves optimal performance for all objective metrics, while controlling the amount of computation. 
Based on the HRSOM dataset and the proposed multi-stage training strategy, we have achieved the best subjective and objective results in the industry and the academia.

\bibliography{refs}

\begin{thebibliography}{10}

\bibitem{yang2013saliency}
Chuan Yang, Lihe Zhang, Huchuan Lu, Xiang Ruan, and Ming-Hsuan Yang,
\newblock ``Saliency detection via graph-based manifold ranking,''
\newblock in {\em Proceedings of the IEEE conference on computer vision and
  pattern recognition}, 2013, pp. 3166--3173.

\bibitem{zhu2014saliency}
Wangjiang Zhu, Shuang Liang, Yichen Wei, and Jian Sun,
\newblock ``Saliency optimization from robust background detection,''
\newblock in {\em Proceedings of the IEEE conference on computer vision and
  pattern recognition}, 2014, pp. 2814--2821.

\bibitem{srivatsa2015salient}
R~Sai Srivatsa and R~Venkatesh Babu,
\newblock ``Salient object detection via objectness measure,''
\newblock in {\em 2015 IEEE international conference on image processing
  (ICIP)}. IEEE, 2015, pp. 4481--4485.

\bibitem{zhang2015minimum}
Jianming Zhang, Stan Sclaroff, Zhe Lin, Xiaohui Shen, Brian Price, and Radomir
  Mech,
\newblock ``Minimum barrier salient object detection at 80 fps,''
\newblock in {\em Proceedings of the IEEE international conference on computer
  vision}, 2015, pp. 1404--1412.

\bibitem{Wang_2021_ICCV}
Xintao Wang, Liangbin Xie, Chao Dong, and Ying Shan,
\newblock ``Real-esrgan: Training real-world blind super-resolution with pure
  synthetic data,''
\newblock in {\em Proceedings of the IEEE/CVF International Conference on
  Computer Vision (ICCV) Workshops}, October 2021, pp. 1905--1914.

\bibitem{cascaded_rgbd_sod}
Jing Zhang, Deng-Ping Fan, Yuchao Dai, Xin Yu, Yiran Zhong, Nick Barnes, and
  Ling Shao,
\newblock ``Rgb-d saliency detection via cascaded mutual information
  minimization,''
\newblock in {\em International Conference on Computer Vision (ICCV)}, 2021.

\bibitem{wang2021raining}
Fanyi Wang and Yihui Zhang,
\newblock ``A de-raining semantic segmentation network for real-time foreground
  segmentation,''
\newblock {\em Journal of Real-Time Image Processing}, vol. 18, pp. 873--887,
  2021.

\bibitem{Qin_2020_PR}
Xuebin Qin, Zichen Zhang, Chenyang Huang, Masood Dehghan, Osmar Zaiane, and
  Martin Jagersand,
\newblock ``U2-net: Going deeper with nested u-structure for salient object
  detection,''
\newblock 2020, vol. 106, p. 107404.

\bibitem{shi2016real}
Wenzhe Shi, Jose Caballero, Ferenc Husz{\'a}r, Johannes Totz, Andrew~P Aitken,
  Rob Bishop, Daniel Rueckert, and Zehan Wang,
\newblock ``Real-time single image and video super-resolution using an
  efficient sub-pixel convolutional neural network,''
\newblock in {\em Proceedings of the IEEE conference on computer vision and
  pattern recognition}, 2016, pp. 1874--1883.

\bibitem{wu2021mobilesal}
Yu-Huan Wu, Yun Liu, Jun Xu, Jia-Wang Bian, Yu-Chao Gu, and Ming-Ming Cheng,
\newblock ``Mobilesal: Extremely efficient rgb-d salient object detection,''
\newblock {\em IEEE Transactions on Pattern Analysis and Machine Intelligence},
  vol. 44, no. 12, pp. 10261--10269, 2022.

\bibitem{lee2021tracer}
Min~Seok Lee, WooSeok Shin, and Sung~Won Han,
\newblock ``Tracer: Extreme attention guided salient object tracing network,''
\newblock {\em arXiv preprint arXiv:2112.07380}, 2021.

\bibitem{Liu2019PoolSal}
Jiang-Jiang Liu, Qibin Hou, Ming-Ming Cheng, Jiashi Feng, and Jianmin Jiang,
\newblock ``A simple pooling-based design for real-time salient object
  detection,''
\newblock in {\em IEEE CVPR}, 2019.

\bibitem{Wu_2022}
Yu-Huan Wu, Yun Liu, Xin Zhan, and Ming-Ming Cheng,
\newblock ``P2t: Pyramid pooling transformer for scene understanding,''
\newblock {\em {IEEE} Transactions on Pattern Analysis and Machine
  Intelligence}, pp. 1--12, 2022.

\bibitem{qin2022}
Xuebin Qin, Hang Dai, Xiaobin Hu, Deng-Ping Fan, Ling Shao, and Luc~Van Gool,
\newblock ``Highly accurate dichotomous image segmentation,''
\newblock in {\em ECCV}, 2022.

\bibitem{qin2021boundary}
Xuebin Qin, Deng-Ping Fan, Chenyang Huang, Cyril Diagne, Zichen Zhang,
  Adri{\`a}~Cabeza Sant'Anna, Albert Suarez, Martin Jagersand, and Ling Shao,
\newblock ``Boundary-aware segmentation network for mobile and web
  applications,''
\newblock {\em arXiv preprint arXiv:2101.04704}, 2021.

\bibitem{kim2022revisiting}
Taehun Kim, Kunhee Kim, Joonyeong Lee, Dongmin Cha, Jiho Lee, and Daijin Kim,
\newblock ``Revisiting image pyramid structure for high resolution salient
  object detection,'' 2022.

\bibitem{ronneberger2015unet}
Olaf Ronneberger, Philipp Fischer, and Thomas Brox,
\newblock ``U-net: Convolutional networks for biomedical image segmentation,''
  2015.

\bibitem{zeng2019highresolution}
Yi~Zeng, Pingping Zhang, Jianming Zhang, Zhe Lin, and Huchuan Lu,
\newblock ``Towards high-resolution salient object detection,'' 2019.

\bibitem{7780454}
F.~Perazzi, J.~Pont-Tuset, B.~McWilliams, L.~Van~Gool, M.~Gross, and
  A.~Sorkine-Hornung,
\newblock ``A benchmark dataset and evaluation methodology for video object
  segmentation,''
\newblock in {\em 2016 IEEE Conference on Computer Vision and Pattern
  Recognition (CVPR)}, 2016, pp. 724--732.

\bibitem{xie2022pyramid}
Chenxi Xie, Changqun Xia, Mingcan Ma, Zhirui Zhao, Xiaowu Chen, and Jia Li,
\newblock ``Pyramid grafting network for one-stage high resolution saliency
  detection,''
\newblock in {\em CVPR}, 2022.

\bibitem{wang2004image}
Zhou Wang, Alan~C Bovik, Hamid~R Sheikh, and Eero~P Simoncelli,
\newblock ``Image quality assessment: from error visibility to structural
  similarity,''
\newblock {\em IEEE transactions on image processing}, vol. 13, no. 4, pp.
  600--612, 2004.

\bibitem{kingma2017adam}
Diederik~P. Kingma and Jimmy Ba,
\newblock ``Adam: A method for stochastic optimization,'' 2017.

\bibitem{8099887}
Lijun Wang, Huchuan Lu, Yifan Wang, Mengyang Feng, Dong Wang, Baocai Yin, and
  Xiang Ruan,
\newblock ``Learning to detect salient objects with image-level supervision,''
\newblock in {\em 2017 IEEE Conference on Computer Vision and Pattern
  Recognition (CVPR)}, 2017, pp. 3796--3805.

\end{thebibliography}

\end{document}